\documentclass[10pt,twocolumn,twoside]{IEEEtran}
\usepackage{amsmath,epsfig}
\usepackage{amssymb}
\usepackage{graphicx}
\usepackage{epstopdf}
\usepackage{url}
\usepackage[noadjust]{cite}

\begin{document}

\title{ Online Second Order Methods for Non-Convex Stochastic Optimizations}
\author{ Xi-Lin Li   
\thanks{San Jose, CA 95132 (E-mail: lixilinx@gmail.com). }	
	}

\maketitle

\begin{abstract}
This paper proposes a family of online second order methods for possibly non-convex stochastic optimizations based on the theory of preconditioned stochastic gradient descent (PSGD), which can be regarded as an enhance stochastic Newton method with the ability to handle gradient noise and non-convexity simultaneously. We have improved the implementations of the original PSGD in several ways, e.g., new forms of preconditioners, more accurate Hessian vector product calculations, and better numerical stability with vanishing or ill-conditioned Hessian, etc.. We also have unrevealed the relationship between feature normalization and PSGD with Kronecker product preconditioners, which explains the excellent performance of Kronecker product preconditioners in deep neural network learning. A software package (\url{https://github.com/lixilinx/psgd_tf}) implemented in Tensorflow is provided to compare variations of stochastic gradient descent (SGD) and PSGD with five different preconditioners on a wide range of benchmark problems with commonly used neural network architectures, e.g., convolutional and recurrent neural networks. Experimental results clearly demonstrate the advantages of PSGD in terms of generalization performance and convergence speed.   
\end{abstract}


\section{Introduction}

Stochastic gradient descent (SGD) and its variations, e.g., SGD with either classic or Nesterov momentum, RMSProp, Adam, adaptive learning rates, etc., are popular in diverse stochastic optimization problems, e.g., machine learning and adaptive signal processing \cite{Widrow85, LeCun98, Sutskever2013, rmsprop, adam, no_more_pesky_mu}. These first order methods are simple and numerically stable, but often suffer from slow convergence and inefficiency in optimizing non-convex models. Off-the-shelf second order methods from convex optimizations, e.g., the quasi-Newton method, conjugate gradient method and truncated Newton method, i.e., the Hessian-free optimization, are attracting more attentions \cite{Martens2012_hessian_free, convex_quasi_newton, stochastic_newton1, stochastic_newton2}, and find many successful applications in stochastic optimizations. Most second order methods require large mini-batch sizes, and have high complexities for large-scale problems. At the same time, searchings for new optimization theories and learning rules are always active, and methods like natural gradient descent, relative gradient descent, equilibrated SGD (ESGD), feature normalization \cite{Cardoso96, Amari96, Equilibrated_mu, selu, batch_normalization}, etc., provide us with great insight into the properties of parameter spaces and cost function surfaces in stochastic optimizations. 

This paper proposes a family of online second order stochastic optimization methods based on the theory of preconditioned SGD (PSGD) \cite{psgd}. Unlike most second order methods, PSGD explicitly considers the gradient noises in stochastic optimizations, and works well with non-convex problems. It adaptively estimates a preconditioner from noisy Hessian vector products with natural or relative gradient descent, and preconditions the stochastic gradient to accelerate convergence. We closely study the performance of five forms of preconditioners, i.e., dense, diagonal, sparse LU decomposition, Kronecker product, and scaling-and-normalization preconditioners. ESGD and feature normalization \cite{batch_normalization, selu} are shown to be closely related to PSGD with specific forms of preconditioners. We consider two different ways to evaluate the Hessian vector product, an important measurement that helps PSGD to adaptively extract the curvature information of cost surfaces. We also recommend three methods to regularize the preconditioner estimation problem when the Hessian becomes ill-conditioned, or numerical errors dominate the Hessian vector product evaluations due to floating point arithmetic. We further provide a software package implemented in Tensorflow\footnote{\url{https://www.tensorflow.org/}} for the comparisons between variations of SGD and PSGD with different preconditioners on a wide range of benchmark problems, which include  synthetic and real world data, and involve most commonly used neural network architectures, e.g., recurrent and convolutional networks. Experimental results suggest that Kronecker product preconditioners, including the scaling-and-normalization one, are particularly suitable for training neural networks since affine maps are extensively used there for feature transformations.               

\section{Background}

\subsection{Notations}

Let us consider the minimization of cost function 
\begin{equation}\label{f_theta}
f(\pmb\theta) = E_z[\ell(\pmb\theta, \pmb z)]
\end{equation}
where $E_z$ takes expectation over random variable $\pmb z$, $\ell$ is a loss function, and $\pmb\theta$ is the model parameter vector to be optimized. For example, in a classification problem, $\ell$ could be the cross entropy loss, $\pmb z$ is a pair of input feature vector and class label, vector $\pmb\theta$ consists of all the trainable parameters in the considered classification model, and $E_z$ takes average over all samples from the training data set. By assuming second order differentiable model and loss, we could approximate $\ell(\pmb\theta, \pmb z)$ as a quadratic function of $\pmb\theta$ within a trust region around $\pmb\theta$, i.e.,
\begin{equation}
\ell(\pmb\theta, \pmb z) = \pmb b_z ^T \pmb \theta + 0.5 \pmb\theta^T \pmb H_z \pmb\theta + a_z
\end{equation} 
where $a_z$ is the sum of approximation errors and constant terms independent of $\pmb\theta$, $\pmb H_z$ is a symmetric matrix, and subscript $z$ in $\pmb b_z$, $\pmb H_z$ and $a_z$ reminds us that these three terms depend on $\pmb z$. Now, we may rewrite (\ref{f_theta}) as 
\begin{equation}
f(\pmb\theta) = \pmb b ^T \pmb \theta + 0.5 \pmb\theta^T \pmb H \pmb\theta + a
\end{equation} 
where $\pmb b = E_z[\pmb b_z]$, $\pmb H = E_z[\pmb H_z]$, and $a = E_z[a_z]$. We do not impose any assumption, e.g., positive definiteness, on $\pmb H$ except for being symmetric. Thus the quadratic surface in the trust region could be non-convex. To simplify our notations, we no longer consider the higher order approximation errors included in $a$, and simply assume that $f(\pmb\theta)$ is a quadratic function of $\pmb\theta$ in the considered trust region.    

\subsection{A Brief Review of PSGD}  

PSGD uses preconditioned stochastic gradient to update $\pmb\theta$ as
\begin{equation}\label{psgd}
\pmb\theta \leftarrow \pmb\theta - \mu  \pmb P \frac{\partial \hat{f}(\pmb\theta)}{\partial\pmb\theta}
\end{equation}
where $0<\mu <1$ is a normalized step size, $\hat{f}(\pmb\theta)$ is an estimate of ${f}(\pmb\theta)$ obtained by replacing expectation with sample average, and $\pmb P$ is a positive definite preconditioner adaptively updated along with $\pmb\theta$. Within the considered trust region, let us write the stochastic gradient, ${\partial \hat{f}(\pmb\theta)}/{\partial\pmb\theta}$, explicitly as
\begin{equation}\label{hat_f_sg}
\frac{\partial \hat{f}(\pmb\theta)}{\partial\pmb\theta} = \hat{\pmb H}\pmb\theta + \hat{\pmb b} 
\end{equation} 
where $\hat{\pmb H}$ and $\hat{\pmb b}$ are estimates of $\pmb H$ and $\pmb b$, respectively. Let $\delta\pmb\theta$ be a random perturbation of $\pmb\theta$, and be small enough such that $\pmb\theta + \delta\pmb\theta$ still resides in the same trust region. Then, (\ref{hat_f_sg}) suggests the following resultant perturbation of stochastic gradient,
\begin{equation}\label{delta_g}
\delta \hat{\pmb g} \overset{\underset{\rm{def}}{}}{=} \frac{\partial \hat{f}(\pmb\theta + \delta\pmb\theta)}{\partial\pmb\theta} - \frac{\partial \hat{f}(\pmb\theta)}{\partial\pmb\theta} = \hat{\pmb H}\delta \pmb\theta = \pmb H \delta\pmb\theta + \pmb\varepsilon
\end{equation}  
where $\pmb\varepsilon$ accounts for the error due to replacing $\hat{\pmb H}$ with $\pmb H$.
Note that by definition, $\delta \hat{\pmb g}$ is a random vector dependent on $\delta\pmb\theta$. PSGD pursues the preconditioner $\pmb P$ via minimizing criterion
\begin{equation}\label{psgd_criterion}
c(\pmb P) = E_{\delta\theta}[\delta \hat{\pmb g}^T \pmb P \delta \hat{\pmb g} + \delta\pmb\theta^T \pmb P^{-1} \delta\pmb\theta]
\end{equation}  
where $E_{\delta\theta}$ takes expectation over $\delta\pmb\theta$. We typically introduce factorization $\pmb P=\pmb Q^T \pmb Q$, and update $\pmb Q$ instead of $\pmb P$ directly as $\pmb Q$ can be efficiently learned with natural or relative gradient descent on the Lie group of nonsingular matrices. Detailed learning rules for $\pmb Q$ with different forms can be found in Appendix A. For our preconditioner estimation problem, relative and natural gradients are equivalent, and further details on these two gradients can be found in \cite{Cardoso96, Amari96}. 
   
Under mild conditions, criterion (\ref{psgd_criterion}) determines a unique positive definite $\pmb P$ \cite{psgd}. The resultant preconditioner is perfect in the sense that it preconditions the stochastic gradient such that
\begin{equation}\label{opt_P}
\pmb P E_{\delta\theta}[\delta \hat{\pmb g} \delta \hat{\pmb g}^T] \pmb P = E_{\delta\theta}[\delta \pmb\theta \delta \pmb\theta^T]
\end{equation} 
which is comparable to relationship 
\begin{equation}\label{newton}
\pmb H^{-1} \delta {\pmb g} \delta {\pmb g}^T \pmb H^{-1} = \delta \pmb\theta \delta \pmb\theta^T
\end{equation} 
where $\delta {\pmb g}$ is the perturbation of noiseless gradient, and we assume that $\pmb H$ is invertible such that $\delta {\pmb g} = \pmb H \pmb\delta\pmb\theta$ can be rewritten as (\ref{newton}). Thus, PSGD can be viewed as an enhanced Newton method that can handle gradient noise and non-convexity at the same time. 

Note that in the presence of gradient noise, the optimal $\pmb P$ and $\pmb P^{-1}$ given by (\ref{opt_P}) are not unbiased estimates of $\pmb H^{-1}$ and $\pmb H$, respectively. Actually, even if $\pmb H$ is positive definite and available,  $\pmb H^{-1}$ may not always be a good preconditioner since it could significantly amplify the gradient noise along the directions of the eigenvectors of $\pmb H$ associated with small eigenvalues, and might lead to divergence. More specifically, \cite{psgd} shows that
\begin{equation}
\pmb H^{-1} E_{\delta\theta}[\delta \hat{\pmb g} \delta \hat{\pmb g}^T] \pmb H^{-1} \ge  E_{\delta\theta}[\delta \pmb\theta \delta \pmb\theta^T]
\end{equation}   
where $\pmb A \ge \pmb B$ means that $\pmb A - \pmb B$ is nonnegative definite.  

\section{Implementations of PSGD}

\subsection{Hessian Vector Product Calculation}

\subsubsection{Approximate Solution}

The original PSGD method relies on (\ref{delta_g}) to calculate the Hessian vector product, $\hat{\pmb H}\delta \pmb\theta$. This numerical differentiation method is simple, and only involves gradient calculations. However, it requires $\delta\pmb\theta$ to be small enough such that $\pmb\theta$ and $\pmb\theta + \delta\pmb\theta$ reside in the same trust region. In practice, numerical error might be an issue when handling small numbers with floating point arithmetic. This concern becomes more grave with the emerging of half precision math in neural network training. Still, the approximate solution is empirically proved to work well, especially with double precision floating point arithmetic, and does not involve any second order derivative calculation. 

\subsubsection{Exact Solution}
An alternative way to calculate the Hessian vector product is via
\begin{equation}\label{exact_Hv}
\frac{\partial}{\partial\pmb\theta} \left\{  \left[\frac{\partial \hat{f}(\pmb\theta)}{\partial\pmb\theta}\right]^T \delta \pmb\theta \right\} = \frac{\partial^2 \hat{f}(\pmb\theta)}{\partial\pmb\theta \partial\pmb\theta^T }  \delta \pmb\theta = \hat{\pmb H}\delta \pmb\theta
\end{equation}    
Here, we no longer require $\delta \pmb\theta$ to be small enough. The above trick is known for a long time \cite{Hv}. However, hand coded second order derivative is error prone even for moderately complicated models.   
Nowadays, this choice becomes attractive due to the wide availability of automatic differentiation softwares, e.g., Tensorflow, Pytorch\footnote{\url{http://pytorch.org/}}, etc.. Note that second and higher order derivatives may not be fully supported in certain softwares, e.g., the latest Tensorflow, version 1.6, does not support second order derivative for its while loop. The exact solution is typically computationally more expensive than the approximate one, although both may have the same order of computational complexity.      

\subsection{Different Forms of Preconditioner}

\subsubsection{Dense Preconditioner}

We call $\pmb P$ a dense preconditioner if it does not have any sparse structure except for being symmetric. A dense preconditioner is practical only for small-scale problems with up to thousands of trainable parameters, since it requires $\mathcal{O}(L^2)$ parameters to represent it, where $L$ is the length of $\pmb\theta$. We are mainly interested in sparse, or limited memory, preconditioners, whose representations only require $\mathcal{O}(L)$ or less parameters. 

\subsubsection{Diagonal Preconditioner} Diagonal preconditioner probably is one of the simplest. From (\ref{psgd_criterion}), we are ready to find the optimal solution as
\begin{equation}\label{jb_p}
\pmb P = {\rm diag}\left( \sqrt{ E_{\delta\theta}[\delta \pmb\theta \odot \delta \pmb\theta] \oslash E_{\delta\theta}[\delta \hat{\pmb g} \odot \delta \hat{\pmb g} ] } \right)
\end{equation}
where $\odot$ and $\oslash$ denote element wise multiplication and division, respectively. For $\delta\pmb\theta$ drawn from standard multivariate normal distribution, $E_{\delta\theta}[\delta \pmb\theta \odot \delta \pmb\theta]$ reduces to a vector with unit entries, and (\ref{jb_p}) gives the equilibration preconditioner in the equilibrated SGD (ESGD) proposed in \cite{Equilibrated_mu}. Thus, ESGD is PSGD with a diagonal preconditioner. The Jacobi preconditioner is not optimal by criterion (\ref{psgd_criterion}), and indeed is observed to show inferior performances in \cite{Equilibrated_mu}. 

\subsubsection{Sparse LU Decomposition Preconditioner}

A sparse LU (SPLU) decomposition preconditioner is given by $\pmb P = \pmb Q^T \pmb Q$, where $\pmb Q=\pmb L\pmb U$, and $\pmb L$ and $\pmb U$ are lower and upper triangular matrices, respectively. To make SPLU preconditioner applicable to large-scale problems, except for the diagonals, only the first $r$ columns of $\pmb L$ and the first $r$ rows of $\pmb U$ can have nonzero entries, where $0\le r \ll L$ is the order of the SPLU preconditioner. 

\subsubsection{Kronecker Product Preconditioner}

A Kronecker product preconditioner is given by $\pmb P = \pmb Q^T \pmb Q$, where $\pmb Q = \ldots\otimes \pmb Q_2\otimes \pmb Q_1$, and $\otimes$ denotes Kronecker product. Kronecker product preconditioners have been previously exploited in \cite{psgd, Martens2015, hess_kron}.      We find that they are particularly suitable for preconditioning gradients of tensor parameters. For example, for a tensor  ${\Theta}$ with shape $(I, J, K)$, we flatten $\Theta$ into a column vector with length $IJK$, and use Kronecker product preconditioner 
\begin{equation}
\pmb P = \pmb P_3 \otimes \pmb P_2 \otimes \pmb P_1
\end{equation}
to have preconditioned gradient $\pmb P {\partial \hat{f}}/{\partial {\rm vec}(\Theta)}$, where $\pmb P_1=\pmb Q_1^T\pmb Q_1$, $\pmb P_2=\pmb Q_2^T\pmb Q_2$, and $\pmb P_3=\pmb Q_3^T\pmb Q_3$ are three positive definite matrices with shapes $(I, I)$, $(J, J)$, and $(K, K)$, respectively. Such a preconditioner only requires $\mathcal{O}(I^2 + J^2 + K^2)$ parameters for its representation, while a dense one requires $\mathcal{O}(I^2J^2K^2)$ parameters. For neural network learning,  ${\Theta}$ is likely to be a matrix parameter, and thus preconditioner of form $\pmb P_2 \otimes \pmb P_1$ should be used. 

\subsubsection{SCaling-And-Normalization (SCAN) Preconditioner}
SCAN preconditioner $\pmb P =  \pmb Q^T \pmb Q  $ is a specific Kronecker product preconditioner specially designed for neural network training, where $\pmb Q = \pmb Q_2 \otimes \pmb Q_1$, $\pmb Q_1$ is a diagonal matrix, and only entries of the diagonal and last column of $\pmb Q_2$ can have nonzero values. As explained in Section IV.B, PSGD with a SCAN preconditioner is equivalent to SGD with normalized input features and scaled output features. It is not difficult to verify that for such sparse $\pmb Q_1$ and $\pmb Q_2$ with positive diagonal entries, matrices with decomposition $\pmb Q_2\otimes \pmb Q_1$ form a Lie group. Hence, natural or relative gradient descent applies to SCAN preconditioner estimation as well.        

It is not possible to enumerate all feasible forms of preconditioners. Except for a few cases, we cannot find closed form solution for the optimal $\pmb P$ with a desired form when given enough pairs of $(\delta\pmb\theta, \delta\hat{\pmb g})$. Hence, it is important to design preconditioners with proper forms such that efficient learning, e.g., natural or relative gradient descent, is available.  Table~I summarizes the ones  we have discussed, and their degrees of freedoms. These simple preconditioners can be used as building blocks for forming larger preconditioners via direct sum and/or Kronecker product operations. All preconditioners obtained in this way can be efficiently learned with natural or relative gradient descent, and such technical details can be found in Appendix A.   

\begin{table}
	\centering
	\caption{ Number of parameters in $\pmb P$ for preconditioning gradient of a matrix parameter with shape $(M, N)$  }\label{table1}
	\begin{tabular}{ll}
		\hline
		Preconditioner & Number of parameters \\ \hline 
		Dense & $0.5(M^2N^2+MN)$ \\  
		SPLU with order $r$ & $2(r+1)MN - r^2 - 2r$ \\  
		Kronecker product & $0.5(M^2+N^2+M+N)$  \\  
		Diagonal & $MN$ \\  
		SCAN & $M+2N-1$ \\ \hline
	\end{tabular}
\end{table} 

\subsection{Regularization of PSGD}

In practice, a second order method might suffer from vanishing second order derivative and ill-conditioned Hessian. The vanishing gradient problem is well known in training deep neural network. Similarly, under certain numerical conditions, the Hessian could be too small to be accurately calculated with floating point arithmetic even if the model and loss are differential everywhere. Non-differential model and loss also could lead to vanishing or ill-conditioned Hessian. For example, second order derivative of the commonly used rectified linear unit (ReLU) is zero almost everywhere, and undefined at the original point. We observe that vanishing or ill-conditioned Hessian could cause numerical difficulties to PSGD, although it does not use the Hessian directly. Here, we propose three remedies to regularize the estimation of preconditioner in PSGD. 

\subsubsection{Traditional Damping}

This method damps the Hessian by adding a diagonal loading term, $\lambda \pmb I$, to it, where $\lambda >0$ is a small number, and $\pmb I$ is the identity matrix. In practice, all we need to do is to replace $(\delta\pmb\theta, \delta\hat{\pmb g})$ with  $(\delta\pmb\theta, \delta\hat{\pmb g} + \lambda \delta\pmb\theta)$ when estimating the preconditioner.

\subsubsection{Non-Convexity Compatible Damping} 

The traditional damping works well only for convex problems. For non-convex optimization, it could make things worse if the Hessian has eigenvalues close to $-\lambda$. In the proposed non-convexity compatible damping, all we need to do is to replace $(\delta\pmb\theta, \delta\hat{\pmb g})$ with  $(\delta\pmb\theta, \delta\hat{\pmb g} + \lambda \delta\pmb\vartheta)$ when estimating the preconditioner, where $\delta\pmb\vartheta$ is a random vector having the same probability density distribution as that of $\delta\pmb\theta$, and independent of $\delta\pmb\theta$. 

\subsubsection{Preconditioned Gradient Clipping}

Gradient clipping is a commonly used trick to stabilize the training of recurrent neural network. We find it useful to stabilize PSGD as well. PSGD with preconditioned gradient clipping is given by
\begin{equation}\label{clipped_psgd}
\pmb\theta \leftarrow \pmb\theta - \frac{\mu \pmb P  {\partial \hat{f}(\pmb\theta)}/{\partial\pmb\theta} }{\max\left(1, \; \|\pmb P  {\partial \hat{f}(\pmb\theta)}/{\partial\pmb\theta}\|/ \Omega \right) } 
\end{equation}
where $\|\cdot\|$ takes norm of a vector, and $\Omega>0$ is a threshold. 

Note that the gradient noises already play a role similar to the non-convexity compatible damping. Thus, we may have no need to use the traditional or the non-convexity compatible damping when the batch size is not too large. Preconditioned gradient clipping is a simple and useful trick for solving problems with badly conditioned Hessians.

\section{Applications to Neural Network Learning}

\subsection{Affine Transformations in Neural Networks} 

Element wise nonlinearity and affine transformation,  
\begin{equation}\label{at}
\pmb y = \pmb \Theta \pmb x
\end{equation}  
are the two main building blocks of most neural networks, where $\pmb\Theta$ is a matrix parameter, and both $\pmb x$ and $\pmb y$ are feature vectors optionally augmented with $1$. Since most neural networks use parameterless nonlinearities, all the trainable parameters are just a list of affine transformation matrices. By assigning a Kronecker product preconditioner to each affine transformation matrix, we are using the direct sum of a list Kronecker product preconditioners as the preconditioner for the whole model parameter vector. Our experiences suggest that this approach provides a good trade off between computational complexities and performance gains. 

It is not difficult to spot out the affine transformations in most commonly used neural networks, e.g., feed forward neural network, vanilla recurrent neural network (RNN), gated recurrent unit (GRU) \cite{gru}, long short-term memory (LSTM) \cite{lstm}, convolutional neural network (CNN) \cite{LeCun98}, etc.. For example, in a two dimensional CNN, the input features may form a three dimensional tensor with shape $(H, W, I)$, and the filter coefficients could form a four dimensional tensor with shape $(H, W, I, O)$, where $H$ is the height of image patch, $W$ is the width of image patch, $I$ is the number of input channels, and $O$ is the number of output channels. 
To rewrite the convolution as an affine transformation, we just need to reshape the filter tensor into a matrix with size $(O, HWI)$, and flatten the input image patch into a column vector with length $HWI$.

\subsection{On the Role of Kronecker Product Preconditioner }

By using Kronecker product preconditioner $\pmb P = \pmb P_2 \otimes \pmb P_1$, the learning rule for $\pmb \Theta$ can be written as
\begin{equation}\label{matrix_p}
\pmb\Theta \leftarrow \pmb\Theta - \mu \pmb P_1 \frac{\partial \hat{f}}{\partial \pmb \Theta } \pmb P_2
\end{equation}
where $\pmb P_1$ and $\pmb P_2$ are two positive definite matrices with proper dimensions. With factorizations $\pmb P_1=\pmb Q_1^T \pmb Q_1$ and $\pmb P_2=\pmb Q_2^T \pmb Q_2$, we can rewrite (\ref{matrix_p}) as
\begin{equation}\label{new_matrix}
\pmb Q_1^{-T} \pmb\Theta \pmb Q_2^{-1} \leftarrow \pmb Q_1^{-T} \pmb\Theta \pmb Q_2^{-1} - \mu \pmb Q_1 \frac{\partial \hat{f}}{\partial \pmb \Theta } \pmb Q_2^T 
\end{equation}
Let us introduce matrix $\pmb\Theta' = \pmb Q_1^{-T} \pmb\Theta \pmb Q_2^{-1} $, and noticing that
\begin{equation}
\frac{\partial \hat{f}}{\partial \pmb \Theta' } = \pmb Q_1 \frac{\partial \hat{f}}{\partial \pmb \Theta } \pmb Q_2^T 
\end{equation}
we can rewrite (\ref{new_matrix}) simply as
\begin{equation}\label{new_sgd}
\pmb\Theta' \leftarrow \pmb\Theta' - \mu   \frac{\partial \hat{f}}{\partial \pmb \Theta' }  
\end{equation}
Correspondingly, the affine transformation in (\ref{at}) is rewritten as $\pmb y' = \pmb\Theta' \pmb x' 
$, where $\pmb y' = \pmb Q_1^{-T} \pmb y$ and $\pmb x' = \pmb Q_2 \pmb x$. Hence, the PSGD in (\ref{matrix_p}) is equivalent to the SGD in (\ref{new_sgd}) with transformed feature vectors $\pmb x'$ and $\pmb y'$. 

We know that feature whitening and normalization could accelerate convergence, and batch normalization and self-normalizing neural networks are such examples \cite{batch_normalization, selu}. Actually, feature normalization can be viewed as PSGD with a specific SCAN preconditioner with constraint $\pmb Q_1=\pmb I$ and a proper $\pmb Q_2$. This fact is best explained by considering the following example with two input features, 
\begin{equation}\label{bn}
\left[
\begin{array}{c}
x_1'   \\
x_2' \\ 
1
\end{array}
\right] = \left[
\begin{array}{ccc}
1/\sigma_1 & 0 & -\nu_1/\sigma_1   \\
0 &  1/\sigma_2 & -\nu_2/\sigma_2 \\
0 &  0 & 1
\end{array}
\right] \left[
\begin{array}{c}
x_1   \\
x_2 \\ 
1
\end{array}
\right]
\end{equation}
where $\nu_i$ and $\sigma_i$ are the mean and standard deviation of $x_i$, respectively. However, we should be aware that explicit input feature normalization is only empirically shown to accelerate convergence, and has little meaning in certain scenarios, e.g., RNN learning where features may not have any stationary distribution. Furthermore, feature normalization cannot normalize the input and output features simultaneously. PSGD provides a more general and principled approach to find the optimal preconditioner, and applies to a much broader range of applications. A SCAN preconditioner does not necessarily  ``normalize'' the input features in the sense of mean removal and variance normalization.   

\section{Experimental Results}

\subsection{Tensorflow Implementation}

Tensorflow is one the most popular machine learning frameworks with automatic differentiation support. We have defined a bunch of benchmark problems with both synthetic and real world data, and implemented SGD, RMSProp, ESGD, and PSGD with five forms of preconditioners. One trick worthy to point out is that in our implementations, we use the  preconditioner from last iteration to precondition the current gradient. In this way, preconditioning gradient and updating preconditioner can be processed in parallel. The original method  in \cite{psgd} updates preconditioner and model parameters sequentially. It may marginally speed up the convergence, but one iteration could take longer wall time.    

To make our comparison results easy to analyze, we try to keep settings simple and straightforward, and do not consider commonly used neural network training tricks like momentum, time varying step size, drop out, batch normalization, etc.. Moreover, tricks like drop out and batch normalization cannot be directly applied to RNN training. Preconditioners of PSGD always are initialized with identity matrix, and updated with a constant normalized step size, $0.01$, and mini-batch size $1$. We always set $r=10$ for the SPLU preconditioner. We independently sample each entry of $\delta\pmb\theta$ from normal distribution with mean $0$ and variance ${\rm eps}$ when (\ref{delta_g}) is used to approximate the Hessian vector product, and mean $0$ variance $1$ when (\ref{exact_Hv}) is used, where ${\rm eps}=2^{-23}$ is the single precision machine epsilon.  We only report the results of PSGD with exact Hessian vector product here since the versions with approximated one typically give similar results. The training loss is smoothed to keep our plots legible. Any further experimental details and results not reported here, e.g., step sizes, preconditioned gradient clipping thresholds, mini-batch sizes, neural network initial guesses, training and testing sample sizes, training loss smoothing factor, etc., can be found in our package at \url{https://github.com/lixilinx/psgd_tf}. 

\subsection{Selected Experimental Results}

\subsubsection{Experiment 1} We consider the addition problem first proposed in \cite{lstm}. A vanilla RNN is trained to predict the mean of two marked real numbers randomly located in a long sequence.   Further details on the addition problem can be found in \cite{lstm} and our implementations. Here, we deliberately use mini-batch size $1$. Fig.~1 summarizes the results. PSGD with any preconditioner outperforms the first order methods. It is clear that PSGD is able to damp the gradient noise and accelerate convergence simultaneously.   

\begin{figure}[h]
	\centering
	\includegraphics[width=\columnwidth]{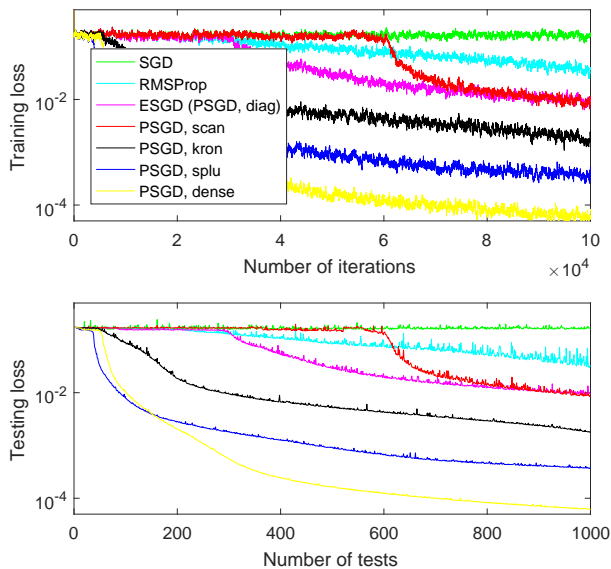}\\
	\caption{ Convergence curves on the addition problem with a standard RNN and mini-batch size $1$. Both the training and testing losses are mean square errors (MSE).   }
\end{figure}

\subsubsection{Experiment 2} Experiment 1 shows that SGD like methods have great difficulties in optimizing certain models. Hence, many thoughtfully designed neural network architectures are proposed to facilitate learning, e.g., LSTM \cite{lstm}, GRU \cite{gru}, residual network \cite{resnet}, etc.. As revealed by its name, LSTM provides the designs for learning tasks requiring long term memories. Still, we find that with first order methods, LSTM completely fails to solve the delayed-XOR benchmark problem proposed in \cite{lstm}. Fig.~2 shows the convergence curves of seven tested methods. Only PSGD with dense, SPLU, Kronecker product and SCAN preconditioners can successfully solve this problem. Bumpy convergence curves suggest that Hessians at local minima is ill-conditioned.  We would like to point out that a vanilla RNN successes to solve this problem when trained with PSGD, and fails when trained with first order methods as well. More details are given in our package. Hence, selecting the right training method is at least as important as choosing a proper model.               

\begin{figure}[h]
	\centering
	\includegraphics[width=\columnwidth]{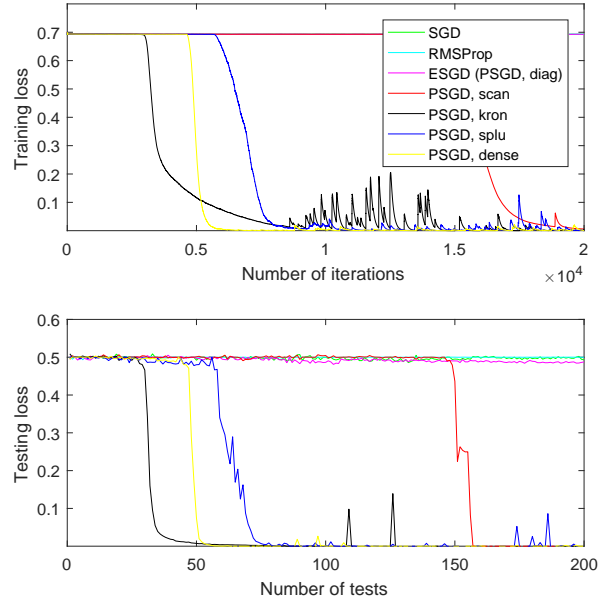}\\
	\caption{ Convergence curves on the delayed-XOR problem with LSTM. Training and testing losses are the logistic loss and classification error rate, respectively.  }
\end{figure}

\subsubsection{Experiment 3} This experiment considers the well known MNIST\footnote{\url{http://yann.lecun.com/exdb/mnist/}} handwritten digits recognition task  using CNN. We do not augment the training data with affine or elastic distorted images. However, we do randomly shift the original training image by $\pm 2$ pixels both horizontally and vertically, and nest the shifted one into a larger, $(32, 32)$, image. A CNN consisting of four convolutional, two average pooling and one fully connected layers is adopted. The traditional nonlinearity, tanh, is used. Fig.~3 shows the convergence curves. PSGD with dense preconditioner is not considered due to its excessively high complexity. We find that PSGD with any preconditioner outperforms the first order methods on the training set.  PSGD with the Kronecker product preconditioner also outperforms the first order methods on the test set, achieving average testing classification error rates about $0.4\%$. Such a performance is impressive as we do not use any complicated data augmentation. 

\begin{figure}[h]
	\centering
	\includegraphics[width=\columnwidth]{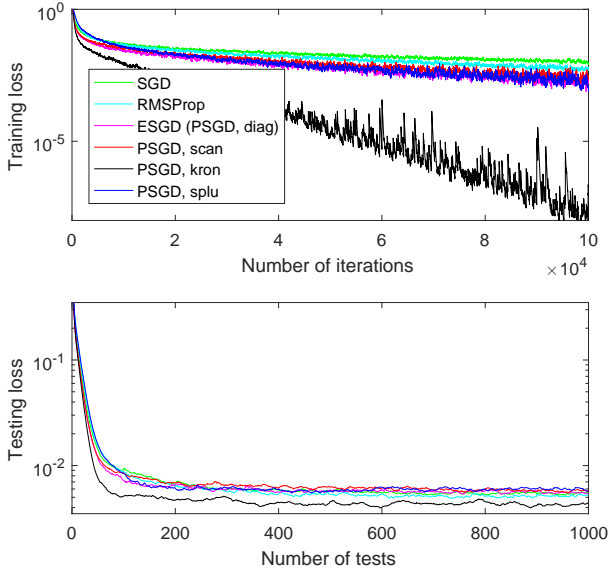}\\
	\caption{ Convergence curves on the MNIST handwritten digits recognition task using CNN. Training and testing losses are cross entropy and classification error rate, respectively. The testing loss is smoothed to make results from different methods more distinguishable.      }
\end{figure}

\subsubsection{Experiment 4}

We consider the CIFAR10\footnote{\url{https://www.cs.toronto.edu/~kriz/cifar.html}} image classification problem using a CNN with four convolutional layers, two fully connected layers, and two max pooling layers. All layers use the leaky ReLU, $\max(0.3x, x)$. Apparently, the model is not second order differentiable everywhere. Nevertheless, PSGD does not use the Hessian, and we find that it works well with such non-differentiable models. Experimental settings are similar to the CIFAR10 classification examples in our package, and here we mainly describe the differences. In this example, we update the preconditioner at the $t$th iteration only if $\mod(t, \max(\lfloor \log_{10}t \rfloor, 1))=0$. The total number of iterations is $500000$. Thus, the average update rate of preconditioner is about once per five iterations. In this way, the average wall time per iteration of PSGD is closely comparable to that of SGD. The Keras ImageDataGenerator is used to augment the training data with setting $0.25$ for the width\_shift\_range, height\_shift\_range and zoom\_range, and default settings for others. Batch size is $100$, step size is $0.0005$ for RMSProp and $0.05$ for other methods. Fig.~4 shows the comparison results. The SPLU preconditioner seems not a good choice here as it has too many parameters, but uses no prior information of the neural network parameter space. PSGD with Kronecker product preconditioners, including the SCAN one, perform the best, achieving average test classification error rates slightly lower than $0.1$. Although the diagonal preconditioner has more parameters than the SCAN one, its performance is just close to that of first order methods.      

\begin{figure}[h]
	\centering
	\includegraphics[width=\columnwidth]{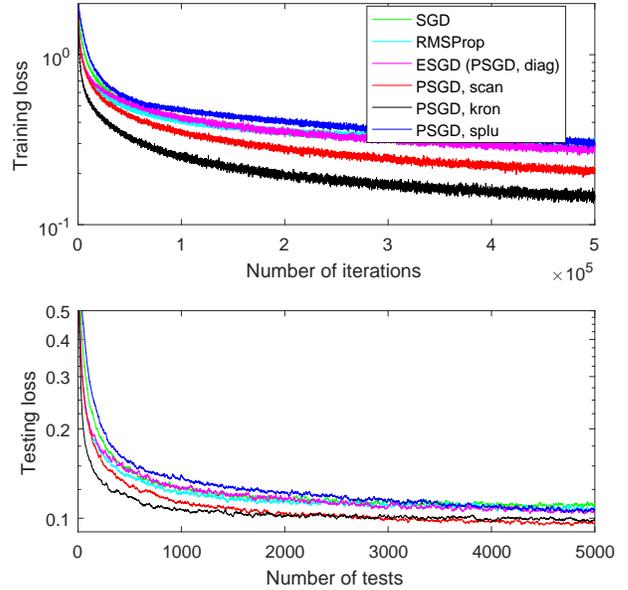}\\
	\caption{ Convergence curves on the CIFAR10 image classification task using CNN. Training and testing losses are cross entropy and classification error rate, respectively. The testing loss is smoothed to make results from different methods more distinguishable.    }
\end{figure}

\subsubsection{Experiment 5} Here, we consider an image autoencoder consisting of three convolution layers for encoding and three deconvolution layers for decoding. Training and testing images are from the CIFAR10 database. Fig.~5 summarizes the results. SGD performs the worst. ESGD and PSGD with SCAN preconditioner give almost identical convergence curves, and outperform RMSProp. PSGD with Kronecker product preconditioner clearly outperforms all the other methods.        

\begin{figure}[h]
	\centering
	\includegraphics[width=\columnwidth]{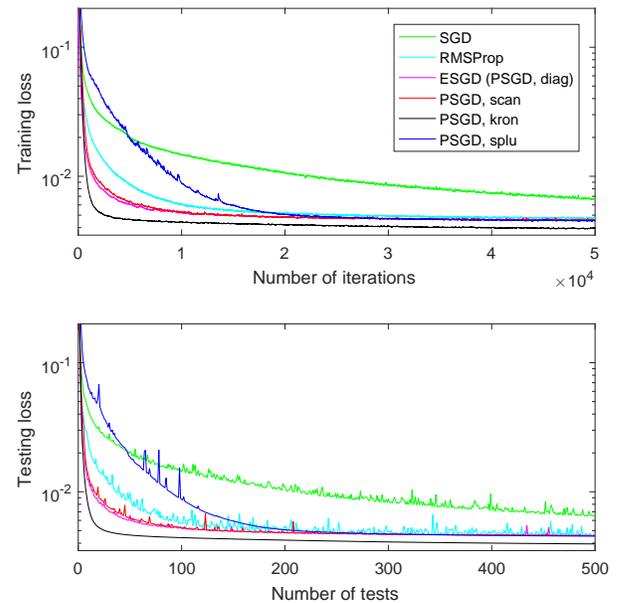}\\
	\caption{ Convergence curves for the image autoencoder trained on CIFAR10 database. Both the training and testing losses are MSEs.   }
\end{figure}

\subsection{Complexities of PSGD}

\subsubsection{Computational Complexities} We consider computational complexity per iteration. Compared with SGD, PSGD comes with three major extra complexities: 
\begin{itemize}
\item C1: evaluation of the Hessian vector product; 
\item C2: preconditioner updating;
\item C3: preconditioned gradient calculation. 
\end{itemize}
C1 typically has the same complexity as SGD \cite{Hv}. Depending on the neural network architectures, complexity of C1 and SGD varies a lot. For the simplest feed forward neural network, SGD has complexity $\mathcal{O}(BMN)$, where $(M, N)$ is the shape of the largest matrix in the model, and $B$ is the mini-batch size. For a vanilla RNN, the complexity rises to $\mathcal{O}(DBMN)$, where $D$ is the back propagation depth. More complicated models may have higher complexities. Complexities of C2 and C3 depend on the form of preconditioner. For a Kronecker product preconditioner, C2 has complexity $\mathcal{O}(\max(M^3, N^3))$, and C3 has complexity $\mathcal{O}(\max(M, N)MN)$. One simple way to reduce the complexities of C2 and C3 is to split those big matrices into smaller ones, and let each smaller matrix keep its own Kronecker product preconditioner. Another practical way is to update the preconditioner less frequently as shown in Experiment 4. Since the curvatures are likely to evolve slower than the gradients, PSGD with skipped preconditioner update often converges as fast as a second order method, and at the same time, has an average wall time per iteration closely comparable to that of SGD.            

\subsubsection{Wall Time Comparisons}

On our machines and with the above benchmark problems, the wall time per iteration of PSGD without skipped preconditioner update typically just doubles that of SGD. This is not astonishing since many parts of PSGD may be processed in parallel. For example, updating preconditioner and preconditioning gradient can be executed in parallel as the preconditioner from last iteration is used to precondition the current gradient. Preconditioners for all the affine transformation matrices in the model can be updated in parallel as well once $(\delta\pmb\theta, \delta\hat{\pmb g})$ is prepared.   

Here, we list the median wall time  per iteration of each method in Experiment 5 running on a GeForce GTX 1080 Ti graphics card: 0.007 s for SGD; 0.008 s for RMSProp; 0.014 s for ESGD; 0.015 s for PSGD with SCAN preconditioner; 0.017 s for PSGD with Kronecker product preconditioner; and 0.017 s for PSGD with sparse LU preconditioner.         

\subsection{Working with Large mini-Batch Sizes}

Using large mini-batch sizes and step sizes could save training time, but also might bring new issues such as poor convergence and over fitting \cite{large_batch}. The gradients become more deterministic with the increase of mini-batch sizes. Without preconditioning and with badly conditioned Hessians, the model parameters will be adapted only along a few directions of the eigenvectors of Hessian associated with large eigenvalues. For many methods, this behavior leads to slow and poor convergences. PSGD seems suffer less from such concerns since the preconditioner is optimized in a way to make $\pmb P \hat{\pmb H}$ has unitary absolute eigenvalues. Hence, the model parameters are updated in a balanced manner in all directions. We have tried mini-batch size $1024$ on our benchmark problems, and observe no meaningful performance loss.  

\subsection{Applications to General Optimization Problems}

The boundary between stochastic and general optimization problems blurs with the increase of mini-batch sizes. Thus, a well designed stochastic optimization method should also work properly on the general mathematical optimization. Preliminary results show that PSGD does perform well on many mathematical optimization benchmark problems. A Rosenbrock function\footnote{\url{https://en.wikipedia.org/wiki/Rosenbrock_function}} minimization demo is included in our package, and PSGD finds the global minimum with about $200$ iterations. Methods like SGD, RMSProp, Adam, batch normalization, etc.,  are apparently not prepared for these problems. Further discussion in this direction strays away from our focus.      

\section{Conclusions}

We have proposed a family of preconditioned stochastic gradient descent (PSGD) methods with approximate and exact Hessian vector products, and with dense, sparse LU decomposition, diagonal, Kronecker product, and SCaling-And-Normalization (SCAN) preconditioners. The approximate Hessian vector product via numerical differentiation is a valid alternative to the more costly exact solution, especially when automatic second order differentiation is unavailable or the exact solution is expensive to obtain. We have shown that equilibrated SGD and feature normalization are closely related to PSGD with specific forms of preconditioners. We have compared PSGD with variations of SGD in several performance indices, e.g., training loss, testing loss, and wall time per iteration, on benchmark problems with different levels of difficulties. These first order methods fail completely on tough benchmark problems with synthetic data, and show inferior performance on problems with real world data. The Kronecker product preconditioners, including the SCAN one, are particularly suitable for training neural networks since affine maps are extensively used there for feature transformations.  A PSGD software package implemented in Tensorflow is available at \url{https://github.com/lixilinx/psgd_tf}.  

\section*{Appendix A: Learning Preconditioners on the Lie Group}

\subsection{Lie Group and Natural/Relative Gradient}

We consider the Lie group of invertible matrices as it is closely related to our preconditioner estimation problem. Assume $\pmb W$ is an invertible matrix. All such invertible matrices with the same dimension as $\pmb W$ form a Lie group. We consider $f(\pmb W)$, a mapping from this Lie group to $\mathbb{R}$. 

In natural gradient descent, the metric tensor depends on $\pmb W$. One example is to define the distance between $\pmb W$ and $\pmb W + d\pmb W$ as 
$${\rm dist}(\pmb W, \pmb W + d\pmb W) = \sqrt{{\rm tr}(d\pmb W \pmb W^{-1}\pmb W^{-T} d\pmb W^T )}$$
where ${\rm tr}$ denotes trace. Intuitively, the parameter space around $\pmb W$ becomes more curved when $\pmb W$ is approaching a singular matrix. With this metric, the natural gradient is given by 
\[ \pmb \nabla_{\rm natural} \pmb W = \frac{\partial f}{\partial \pmb W} \pmb W^T \]  
In relative gradient descent, we assume $d\pmb W = \mathcal{E} \pmb W$, and it can be shown that
\[ \pmb \nabla_{\rm relative}\pmb W =  \left. \frac{\partial {f}(\pmb W + \mathcal{E} \pmb W)}{\partial \mathcal{E}} \right|_{\mathcal{E} = \pmb 0} = \frac{\partial f}{\partial \pmb W} \pmb W^T \]
Hence, the natural and relative gradients have the same form.  

It is also possible to choose other metrics in the natural gradient. For example, natural gradient with metric ${\rm tr}(d\pmb W^T \pmb W^{-T}  \pmb W^{-1}  d\pmb W)$ and relative gradient with form $d\pmb W = \pmb W \mathcal{E} $ lead to another form of gradient,
\[ \pmb \nabla \pmb W =    \pmb W^T \frac{\partial f}{\partial \pmb W}  \]
For a specific problem, we should choose the natural or relative gradient with a form that can simplify the gradient descent learning rule. For example, for mapping $f(\pmb W) = \det(\pmb W)$, both the above considered metrics lead to natural gradient $\pmb I$, while metric ${\rm tr}(d\pmb W \pmb W^{-T}  \pmb W^{-1}  d\pmb W^T )$ leads to natural gradient $\pmb W^{-T}\pmb W$. Clearly, the previous form is preferred due to its simplicity. For preconditioner learning, we are most interested in groups that have sparse representations. 

\subsection{Learning on the Group of Triangular Matrices}

It is straightforward to show that upper or lower triangular matrices with positive diagonals form a Lie group. Natural and relative gradients may have the same form on this group as well. For learning rule
\[ \pmb W \leftarrow \pmb W  - \mu \pmb \nabla \pmb W \, \pmb W, \; {\rm or}\; \pmb W \leftarrow \pmb W  - \mu  \pmb W \; \pmb \nabla \pmb W \]
it is convenient to choose the following normalized step size,
\[ \mu = \frac{\mu_0}{\| \pmb \nabla \pmb W \|} \]
where $0< \mu_0<1$, and $\| \pmb \nabla \pmb W \|$ is a matrix norm of $\pmb \nabla \pmb W$.  In our implementations, we simply use the max norm. It is clear that the new $\pmb W$ still resides on the same Lie group with this normalized step size.  

\subsection{Natural/Relative Gradients for $\pmb Q$ with Different Forms}

\subsubsection{Arbitrary Invertible $\pmb Q$}

We let $ d\pmb Q = \mathcal{E} \pmb Q $. Then the relative gradient of criterion (\ref{psgd_criterion}) with respect to $\pmb Q$ can be shown to be
\[ \pmb \nabla \pmb Q = 2E[ \pmb Q\delta \hat{\pmb g} \delta \hat{\pmb g}^T \pmb Q^T - \pmb Q^{-T}\delta \pmb \theta \delta \pmb \theta^T \pmb Q^{-1} ] \]
In practice, it is convenient to constrain $\pmb Q$ to be a triangular matrix such that $\pmb Q^{-T}\delta \pmb \theta$ can be efficiently calculated by back substitution. For triangular $\pmb Q$ with positive diagonals, $\pmb P=\pmb Q^T\pmb Q$ is known to be the Cholesky factorization of $\pmb P$.

\subsubsection{$\pmb Q$ with Factorization $\pmb Q = \pmb Q_2\otimes \pmb Q_1$}

We let
\[ d \pmb Q_1=\mathcal{E}_1 \pmb Q_1, \;  d \pmb Q_2 = \mathcal{E}_2 \pmb Q_2 \]
Then, the relative gradients can be shown to be
\begin{eqnarray*}
	\pmb A & =& \pmb Q_1 \delta \hat{\pmb G} \pmb Q_2^T \\
	\pmb B &=& \pmb Q_2^{-T} \delta\pmb \Theta^T \pmb Q_1^{-1} \\
	 \pmb \nabla \pmb Q_1  & = & 2 E [ \pmb A\pmb A^T - \pmb B^T\pmb B  ]  \\
 \pmb \nabla \pmb Q_2   & = & 2 E [ \pmb A^T\pmb A - \pmb B\pmb B^T  ] 
\end{eqnarray*}
where $\delta \hat{\pmb G}$ is the $\delta \hat{\pmb g}$ rewritten in matrix form. Again, by constraining $\pmb Q_1$ and $\pmb Q_2$ to be triangular matrices,  $\pmb B$ can be efficiently calculated with back substitution.

\subsubsection{$\pmb Q$ with Direct Sum Decomposition} 

For $\pmb Q = \pmb Q_1 \oplus \pmb Q_2$, we can update $\pmb Q_1$ and $\pmb Q_2$ separately as they are orthogonal, where $\oplus$ denotes direct sum. 

\subsubsection{ $\pmb Q$ with LU Decomposition }

We assume $\pmb Q$ has LU decomposition $ \pmb Q = \pmb L\pmb U $, where
\[ \pmb L = \left[
\begin{array}{cc}
\pmb L_1 &      \\
\pmb L_2 &  \pmb L_3
\end{array}
\right] , \;  \pmb U = \left[
\begin{array}{cc}
\pmb U_1 & \pmb U_2    \\
  &  \pmb U_3
\end{array}
\right] \]
are lower and upper triangular matrices, respectively. When $\pmb L_3$ and $\pmb U_3$ are diagonal matrices, this LU decomposition is sparse.   

Let $ dL = \mathcal{E}_l \pmb L$ and $ dU = \pmb U \mathcal{E}_u  $. Then, the relative gradients of criterion (\ref{psgd_criterion}) with respect to $\pmb  L$ and $\pmb U$ can be shown to be
\[ \pmb \nabla \pmb L = 2E[ \pmb Q\delta \hat{\pmb g} \delta \hat{\pmb g}^T \pmb Q^T - \pmb Q^{-T}\delta \pmb \theta \delta \pmb \theta^T \pmb Q^{-1} ] \]
\[ \pmb \nabla \pmb U = 2E[ \pmb P\delta \hat{\pmb g} \delta \hat{\pmb g}^T  - \delta \pmb \theta \delta \pmb \theta^T \pmb P^{-1}  ] \]
For sparse $\pmb L$ and $\pmb U$, we let $\pmb L_3 = {\rm diag}(\pmb l_3)$ and $\pmb U_3 = {\rm diag}(\pmb u_3)$. Then, with relationships
\[ \pmb L^{-1} = \left[
\begin{array}{cc}
\pmb L_1^{-1} &     \\
- {\rm diag}(\pmb l_3)^{-1}  \pmb L_2 \pmb L_1^{-1} &  {\rm diag}(\pmb l_3)^{-1} 
\end{array}
\right] \]       
\[ \pmb U^{-1} = \left[
\begin{array}{cc}
\pmb U_1^{-1} & -\pmb U_1^{-1} \pmb U_2  {\rm diag}(\pmb u_3)^{-1}    \\
  &  {\rm diag}(\pmb u_3)^{-1} 
\end{array}
\right] \]
it is trivial to inverse $\pmb Q$ and $\pmb P$ as long as $\pmb L_1$ and $\pmb U_1$ are small triangular matrices.

\end{document}